\documentclass[runningheads]{llncs}
\usepackage[T1]{fontenc}
\usepackage{graphicx}
\usepackage{amsfonts}
\usepackage{adjustbox}
\usepackage{ulem}

\usepackage{multirow}
\usepackage{tabularray}
\usepackage{xspace}
\usepackage{amsmath}
\usepackage[table]{xcolor}
\usepackage{subfigure}
\usepackage{booktabs}
\usepackage{graphicx}
\usepackage{capt-of}
\usepackage{marvosym}

\newcommand{\ie}{\textit{i}.\textit{e}.}
\newcommand{\eg}{\textit{e}.\textit{g}.}

\begin{document}
\title{Revisiting Cross-Domain Problem for LiDAR-based 3D Object Detection}
%
%

\author{Ruixiao Zhang\inst{1}\textsuperscript{(\Letter)} \and
Juheon Lee\inst{2} \and
Xiaohao Cai\inst{1} \and
Adam Prugel-Bennett\inst{1}}
\authorrunning{R. Zhang et al.}
%
\institute{ECS, University of Southampton, Southampton, UK \\
\email{\{rz6u20,X.Cai\}@soton.ac.uk, apb@ecs.soton.ac.uk} \and
Meta, USA \\
\email{juheon.lee.626@gmail.com}}
\maketitle              
\begin{abstract}
Deep learning models such as convolutional neural networks and transformers have been widely applied to solve 3D object detection problems in the domain of autonomous driving. While existing models have achieved outstanding performance on most open benchmarks, the generalization ability of these deep networks is still in doubt. To adapt models to other domains including different cities, countries, and weather, retraining with the target domain data is currently necessary, which hinders the wide application of autonomous driving. In this paper, we deeply analyze the cross-domain performance of the state-of-the-art models. We observe that most models will overfit the training domains and it is challenging to adapt them to other domains directly. Existing domain adaptation methods for 3D object detection problems are actually shifting the models' knowledge domain instead of improving their generalization ability. We then propose additional evaluation metrics -- the side-view and front-view AP -- to better analyze the core issues of the methods' heavy drops in accuracy levels. By using the proposed metrics and further evaluating the cross-domain performance in each dimension, we conclude that the overfitting problem happens more obviously on the front-view surface and the width dimension which usually faces the sensor and has more 3D points surrounding it. Meanwhile, our experiments indicate that the density of the point cloud data also significantly influences the models' cross-domain performance.

\end{abstract}

\begin{figure}[htbp]
\centering
\includegraphics[width=1.0\linewidth]
{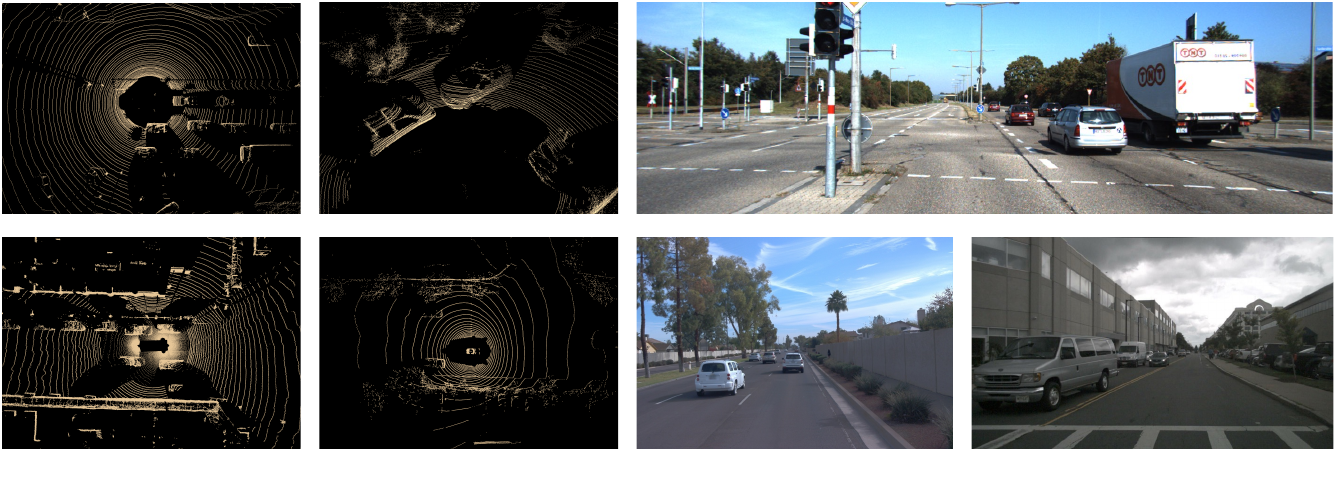}
\put(-320,64){\scriptsize KITTI}
\put(-245,64){\scriptsize KITTI details}
\put(-105,64){\scriptsize KITTI}
\put(-320,3){\scriptsize Waymo}
\put(-240,3){\scriptsize nuScenes}
\put(-155,3){\scriptsize Waymo}
\put(-64,3){\scriptsize nuScenes}
\vspace{-0.15in}
\caption{\small LiDAR point cloud and image data from three datasets: KITTI~\cite{6248074}, Waymo~\cite{Sun_2020_CVPR} and nuScenes~\cite{9156412}. Point cloud density and image shapes are different due to different sensor equipment. For the car objects close to the sensors, a large number of points are collected and most shapes are clearly visible. While for those away from the sensors, only a few points are collected and it is difficult to estimate the dimensions. } 
\label{fig:pcd_example}

\vspace{-0.20in}
\end{figure}
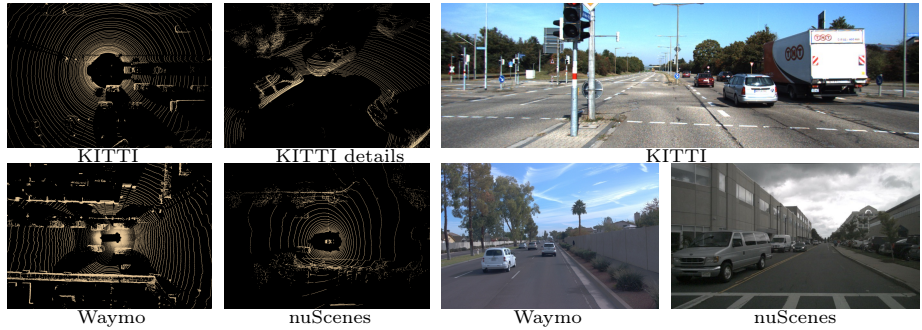

\vspace{-0.25in}
\section{Introduction}
\label{sec:intro}
\vspace{-0.05in}

3D object detection aims to localize and categorize different types of objects in specific 3D space described by 3D sensor data (\eg, LiDAR point clouds). Recently, the application of this technology has achieved significant improvement due to the development of deep neural networks, especially in the field of autonomous driving. Current 3D object detection methods mainly focus on specific datasets, \ie, models will be trained and tested independently on a specific dataset. In doing so, a number of models achieved high performances on public benchmarks including nuScenes~\cite{9156412}, Waymo~\cite{Sun_2020_CVPR}, and KITTI~\cite{6248074} (see \eg, Figure~\ref{fig:pcd_example}).  However, if the evaluation on a new dataset is needed, in most cases, the training on the new dataset as well as modifications of some training hyper-parameters are necessary. In other words, it is hard for models trained on one dataset to adapt directly to another. These domain shifts may arise from different sensor types, weather conditions~\cite{9156543} and object sizes~\cite{9578132} between different datasets or domains. This domain adaptation problem is therefore a big challenge for real-world applications of existing 3D object detection methods, whose retraining steps can be very slow and resource-consuming. It is thus significant to understand the reasons for this cross-domain performance drop and propose efficient methods to raise the cross-domain performance to the same level as within-domain tasks.  

The main factors influencing cross-domain performance can be divided into the models and the domains (datasets). How the domains influence cross-domain performances has been investigated in~\cite{9156543}. By comparing the performance of two models~\cite{8954080,8578896} on several different datasets,  the work in \cite{9156543} has proved that the difference in car size across geographic locations is one of the main challenges for domain adaptation problems. It is then natural to ask: Will there also be any crucial factors influencing the cross-domain performance on the side of models? Therefore, in this paper, our first goal is to investigate the cross-domain performance of existing 3D object detection models with different inputs and structures. We select representative LiDAR-only and multi-modal (\ie, LiDAR + RGB Image) methods including PV-RCNN~\cite{Shi_2020_CVPR}, SECOND~\cite{s18103337} and TransFusion~\cite{Bai_2022_CVPR}, some of which are based on 3D convolutional neural networks (CNNs) and the others are based on transformers. Based on our results on three different datasets, \ie, KITTI~\cite{6248074}, Waymo~\cite{Sun_2020_CVPR} and nuScenes~\cite{9156412}, we find that all the tested methods similarly fail on the cross-domain tasks, no matter they are based on CNNs or transformers. A more interesting fact is that multi-modal methods achieve poorer performance on some tasks than LiDAR-only methods, even though more data information is taken.

There have been some generic training methods trying to overcome the domain adaptation problem, which can also be considered explorations on the side of models. By reproducing and analyzing one of the state-of-the-art (SOTA) methods, ST3D~\cite{9578132}, we surprisingly find that ideas in this field have still limited performance, and more deep ideas are still needed to better solve this problem.

Meanwhile, we notice that current evaluation metrics mainly focus on the average precision (AP) of 3D detection predictions and bird’s-eye view (BEV) predictions, which sometimes are not sufficient to evaluate the performance difference between methods especially when different domains are involved. Therefore, we propose two additional evaluation metrics -- side-view AP and front-view AP -- to make fairer and more comprehensive comparisons. Inspired by the conclusion in~\cite{9156543} that car size difference is the core reason for failure in cross-domain tasks, we evaluate the side-view AP and front-view AP of existing methods, as well as the overlaps between predictions and ground truth in every single dimension. 

Our experiments show that the performance under the side-view and front-view metrics is similar, and the prediction accuracy in object length (\ie, depth) is even higher than that in width. By further analyzing the absolute error in each dimension, we prove that the higher overlap accuracy in length is actually because length is usually much higher than width, and the absolute error in length and width are similar. This conclusion might be contradictory to our common sense at first glance, since when we look at the LiDAR point cloud data such as shown in Figure~\ref{fig:pcd_example}, for most objects, only sides facing or close to the LiDAR sensor can be completely captured. 
The results also illustrate that the overfitting problem of models makes them not care about the completeness of objects' point cloud data in the target domain, but instead make similar predictions as in the source domain where they are trained.

We, therefore, also suggest more investigations in the evaluation methods for 3D object detection, especially in cross-domain tasks, where the current measurements usually focus on the overall overlap of the entire bounding boxes and ignore the different degrees of influence by different dimensions.
To summarize, our main contributions are threefold:
\begin{itemize}
\item We analyze the cross-domain performance of representative models of different structures. Our results show that most existing models overfit to the source domain and cannot directly perform well on other domains, no matter whether based on CNNs or transformers. We also suggest that multi-modal methods are harder to be adapted to new domains due to the inconsistency of data and calibrations.

\item We analyze one of the SOTA self-training methods ST3D for domain adaptation. Our results reveal a serious problem, \ie, the self-training method ST3D actually shifts the knowledge distribution contained in the model to the new domain, which cannot improve the models' generalization ability, but instead reduces the models' detection ability on the source domain.

\item We propose two additional evaluation metrics -- the side-view AP and front-view AP -- to evaluate the models' cross-domain performance more comprehensively and locate the errors more accurately. {Our results illustrate a surprising phenomenon, \ie, although the incomplete point cloud data occurs more in the length dimension than the width dimension of objects due to occlusion, the cross-domain performance under the side-view and front-view metrics is similar. This suggests that the poor cross-domain detection ability of existing models/methods is not directly related to the objects' incomplete point cloud data, but more related to the overfitting problem caused by the model structures and training strategy.}
\end{itemize}

\vspace{-0.15in}
\section{Related Work}
\label{sec:related}
\vspace{-0.05in}
\noindent\textbf{3D object detection with point clouds.} The common way of representing real-world 3D space is using LiDAR point cloud data, which uses 3D points to record the 3D environment. Although it benefits from accurate point locations, the main challenge for LiDAR-based 3D object detection methods is finding the best way to process the point cloud data. CNNs have been widely used in 2D object detection problems; however, due to the sparsity and spatial disorder of the point cloud data, they cannot be directly fed into the CNNs. Therefore, current LiDAR-based methods either transform the point clouds into spatial invariant formats to use CNN models or propose new methods that can learn features directly from the 3D points. VoxelNet~\cite{8578570} and SECOND~\cite{s18103337} encoded point clouds into voxels so that features can be extracted by CNNs designed for 3D inputs. MV3D~\cite{8100174} projected point clouds into 2D spaces (\ie, the front view and bird's-eye view) and used 2D CNNs to extract features. PointRCNN~\cite{8954080} applied PointNet++~\cite{NIPS2017_d8bf84be} to obtain 3D point-wise features and directly learn the 3D proposals from the points. PV-RCNN~\cite{Shi_2020_CVPR} voxelized the point cloud data first and then used key-point-wise features to keep more semantic features, in which the combination of point-based and voxel-based methods greatly improves the performance with acceptable computation cost.

\noindent\textbf{3D object detection with images.} While using 3D data (\eg, point clouds) to detect 3D objects is a reasonable way, there is still a lot of interest in directly adapting 2D methods to 3D fields, which mainly take 2D images as the input data. The main challenge of applying 2D models to 3D problems is how to estimate the depth information of the relevant 3D scene, which cannot be directly obtained from the image data. In \cite{MonoRCNN_ICCV21}, the physical and visual height of objects was used to estimate the depth. The work in \cite{CaDDN} estimated the depth of every pixel from the images to get 3D voxel features and projects the voxel features into the bird's-eye view for the final 3D detection. Although there have been many trials on image-only 3D object detection problems, the lack of rich spatial information still results in a large gap in the performance compared with that predicted by models using the point cloud data.

\noindent\textbf{3D object detection with multi-modal inputs.} Since images contain richer semantic information and point clouds contain more spatial information, it is natural to explore the possibility of fusing these two types of data. Based on the position of the fusion step in models, existing works on multi-modal inputs can be divided into three categories. Earlier works~\cite{8578200,8813895} mainly focused on proposal-level and result-level fusions, in which the models fused the proposals or final predictions obtained from the point cloud channel and the image channel separately. Since the proposals or predictions are based only on one type of data for each channel, both of them suffer from the disadvantages of specific data types and therefore the fusions cannot achieve significantly better results than just using the point cloud data. Afterward, the proposal of PointPainting~\cite{9156790} proved that fusing these two data types in an early step, \ie, the point-level fusion, can greatly improve performance. TransFusion~\cite{Bai_2022_CVPR} pointed out the limitations of hard association in previous fusion models and proposed a soft association between image features and point cloud features, which greatly increases the performance to a higher level than point-cloud-only methods. In this paper, we will focus on both the LiDAR-only methods and the multi-modal methods.

\noindent\textbf{Domain adaptation.} Domain adaptation has been widely used in 2D object detection~\cite{9008383,8953674} and 2D semantic segmentation~\cite{8578921,Huang_2018_ECCV}. However, there are only a few approaches specifically designed for 3D object detection. ST3D~\cite{9578132} and ST3D++~\cite{yang2021st3d++} used self-training algorithms to generate pseudo labels and train models on the target domain without ground truth. In \cite{wei2022lidar}, distillation methods were proposed for LiDAR point clouds to overcome the beam difference between datasets. The work in \cite{9156543} normalized the object size of different datasets based on additional prior knowledge. Our experiments below show that these existing domain adaptation methods still have limited and unstable performance.

\vspace{-0.10in}
\section{Datasets}
\label{sec:data}


\noindent\textbf{KITTI.} The KITTI object detection dataset~\cite{6248074} is one of the most popular datasets in outdoor 3D object detection tasks. It contains 7,481 training samples and 7,518 test samples. For each sample, KITTI provides its point cloud data with a 64-beam Velodyne LiDAR sensor and its image data with stereo cameras. Following existing works~\cite{nips15chen,Shi_2020_CVPR}, the training set is further separated into 3,712 and 3,769 samples as the training and validation sets, respectively.

\noindent\textbf{nuScenes.} The nuScenes dataset~\cite{9156412} contains 28,130 training samples and 6,019 validation samples. Following~\cite{9156543}, we treat the validation set as the test set and re-split the training set into the training and validation sets. Furthermore, since our experiments mainly focus on car detection, we filter out the samples that do not contain any car objects. Finally, there are 8,614 training samples and 2,395 validation samples. For each sample, nuScenes provides its point cloud data with a 32-beam LiDAR and its image data with five cameras for different angles.

\noindent\textbf{Waymo.} The Waymo open dataset~\cite{Sun_2020_CVPR} contains 122,000 training, 30,407 validation, and 40,077 test samples. It is much larger than the other two above-mentioned datasets. Following existing works, we sub-sample the training and validation sets into 7,905 and 2,000 samples. It should be noted that subsampling the dataset by such ratios will not significantly influence the final performance, since Waymo collects the data as continuous frames and models in our experiments do not consider this time-related information.

\noindent\textbf{Data integration.} Since most existing 3D object detection methods focus on the performance within each specific domain/dataset, they often fine-tune the models for different datasets independently and ignore the influence of gaps between them. However, to investigate the cross-domain performance of these models, we must find a way to merge these datasets. We note that the following differences between datasets have a significant influence on the cross-domain experiments: (i) the point cloud range; (ii) the origin of coordinates; and (iii) the unit for preprocessing the point cloud data, such as voxel sizes in voxel-based methods. Following the ideas in~\cite{9156543,9578132}, some preprocessing methods are therefore adopted. For all datasets, we set the point cloud range to $[-75.2, -75.2, -2, 75.2, 75.2, 4]$ m and shift the whole point cloud space of different datasets so that the X-Y plane always coincides with the horizontal plane; following~\cite{9578132}, we set the voxel size of all voxel-based methods to $(0.1, 0.1, 0.15)$~m.

\vspace{-0.10in}
\section{Setup and Metrics}
\vspace{-0.05in}

\subsection{Setup of 3D Object Detection Methods}
 To better investigate the influence of model structures on cross-domain performance, we train and test the following models on KITTI, Waymo, and nuScenes datasets. We first evaluate two representative LiDAR-only methods, \ie, PV-RCNN~\cite{Shi_2020_CVPR} and SECOND~\cite{s18103337}, which are based on 3D CNNs. Afterward, we test the performance of TransFusion~\cite{bai2021pointdsc}, a transformer-based method that can take both the LiDAR point clouds and RGB images as the input, to analyze the influence of adding images in cross-domain tasks. We also compare the results of TransFusion-L (\ie, the Transfusion that only takes LiDAR point clouds as the input) with PV-RCNN and SECOND. Last, we apply ST3D~\cite{9578132}, a SOTA self-training method for domain adaptation problems in 3D object detection, to PV-RCNN. Following the experiments in ST3D~\cite{9578132}, we equip the SECOND model with an extra IoU head for better performance. We first train PV-RCNN, SECOND-IoU, and TransFusion-L using the OpenPCDet~\cite{openpcdet2020} toolbox with suggested numbers of epochs and learning rates. Since OpenPCDet only supports LiDAR-only models, we use another toolbox, MMDetection3D~\cite{mmdet3d2020}, for the comparison of TransFusion-L and TransFusion-LC. 
 
 We train the LiDAR-only model for 40 epochs with learning rate $5\times10^{-5}$ and batch size 8, and further train with images for 20 epochs with the same learning rate and batch size. We train ST3D with PV-RCNN using the OpenPCDet~\cite{openpcdet2020} toolbox. As guided by the original work, we first train the model on the source domain and adapt random object scaling (ROS) to it. Afterward, we train the model with the ST3D method and evaluate the performance on the target domain. Following other works~\cite{bai2021pointdsc,9578132} based on MMDetection3D and OpenPCDet, we adopt random horizontal flip, rotation and scale transforms during the training process. All the models are trained on RTX 8000.


\begin{table*}[ht]
\scriptsize
\caption{Performance of 3D object detection models within and across multiple datasets (evaluated on the validation set). Three representative models are selected for CNN (point-based~\cite{8954080} \& voxel-based~\cite{s18103337}) and transformer methods~\cite{Bai_2022_CVPR}. We report the  AP (average precision) of the \textit{Car} category objects in the format of BEV / 3D with IoU threshold set to 0.7 following the KITTI benchmark. Following \cite{9156543}, we replace the $40, 25, 25$ pixel thresholds on 2D box height with $30, 70, 70$ meters on object depth to better evaluate the performance on Waymo and nuScenes. Results of three within-domain and cross-domain tasks are reported. The results show significant drops when directly adapting models to new domains.}
\begin{center}
\setlength\tabcolsep{6pt}
\resizebox{0.98\textwidth}{!}{
\begin{tabular}{c c c c c}
\toprule
Tasks & Metrics & PV-RCNN & SECOND-IoU & TransFusion-L \\ 
\midrule
\multirow{3}{*}{KITTI $\rightarrow$ KITTI}     
     & Easy      & 95.0 / 91.2 & 94.0 / 89.4 & 90.9 / 82.6 \\
     & Moderate  & 81.7 / 70.5 & 76.5 / 64.5 & 73.7 / 59.6 \\
     & Hard      & 81.4 / 69.0 & 76.2 / 62.7 & 73.0 / 57.6 \\
\midrule
\multirow{3}{*}{nuScenes $\rightarrow$ nuScenes}     
     & Easy      & 57.0 / 41.3 & 55.0 / 37.3 & 54.0 / 33.0 \\
     & Moderate  & 51.7 / 37.3 & 49.9 / 32.9 & 48.8 / 29.3 \\
     & Hard      & 51.7 / 37.3 & 49.9 / 32.9 & 48.8 / 29.3 \\
\midrule
\multirow{3}{*}{nuScenes $\rightarrow$ KITTI}     
     & Easy      & 80.9 / 38.1 & 56.4 / 14.7 & 54.6 / 14.7 \\
     & Moderate  & 63.0 / 25.3 & 37.2 / 8.5  & 38.2 / 9.9  \\
     & Hard      & 62.0 / 24.6 & 36.3 / 7.5  & 38.4 / 10.0 \\
\midrule
\multirow{3}{*}{Waymo $\rightarrow$ Waymo}     
     & Easy      & 70.6 / 62.5 & 68.2 / 59.3 & 68.7 / 57.1 \\
     & Moderate  & 64.5 / 53.9 & 62.3 / 50.3 & 62.8 / 50.3 \\
     & Hard      & 64.5 / 53.9 & 62.3 / 50.3 & 62.8 / 50.3 \\
\midrule
\multirow{3}{*}{Waymo $\rightarrow$ nuScenes}     
     & Easy      & 37.0 / 24.5 & 32.6 / 20.7 & 34.9 / 19.1 \\
     & Moderate  & 32.7 / 21.2 & 29.2 / 18.4 & 31.4 / 16.9 \\
     & Hard      & 32.7 / 21.2 & 29.2 / 18.4 & 31.4 / 16.9 \\
\midrule
\multirow{3}{*}{Waymo $\rightarrow$ KITTI}     
     & Easy      & 75.4 / 25.1 & 66.3 / 25.6 & 73.6 / 35.3 \\
     & Moderate  & 55.9 / 18.9 & 48.1 / 17.3 & 55.6 / 26.9 \\
     & Hard      & 53.2 / 17.8 & 45.2 / 15.0 & 54.7 / 26.0 \\

\bottomrule
\end{tabular}
}
\end{center}
\label{tab:LiDAR-only}
\vspace{-0.15in}
\end{table*}

\vspace{-0.05in}
\subsection{Metrics}\label{metrics}
Most existing methods follow KITTI to evaluate the detection performance in 3D and the BEV by the AP. As concluded in~\cite{9156543}, the car size difference is one of the main idiosyncrasies that account for the performance gap between within-domain and cross-domain tasks. We therefore further ask the following question: Does the gap fairly come from the three dimensions or are there specific dimensions responsible for the majority of the gap? To explore this problem, we propose two additional evaluation metrics, \ie, the side-view AP and the front-view AP, and combine them with the 3D and BEV AP to understand the prediction results better. 

\begin{figure}[htbp]
\vspace{-0.15in}
\noindent
\begin{minipage}[t]{0.46\textwidth} 
  \centering
  \captionof{figure}{\small Definition of the side-view and the front-view AP. 
  The red and blue boxes denote the ground truth and predictions. We project not only the related side to the front/side 2D plane but also consider the other sides that actually can be seen in the related view. For example, the left side is also considered when making a projection into the front view.} \vspace{0.05in}
  \includegraphics[width=\linewidth]{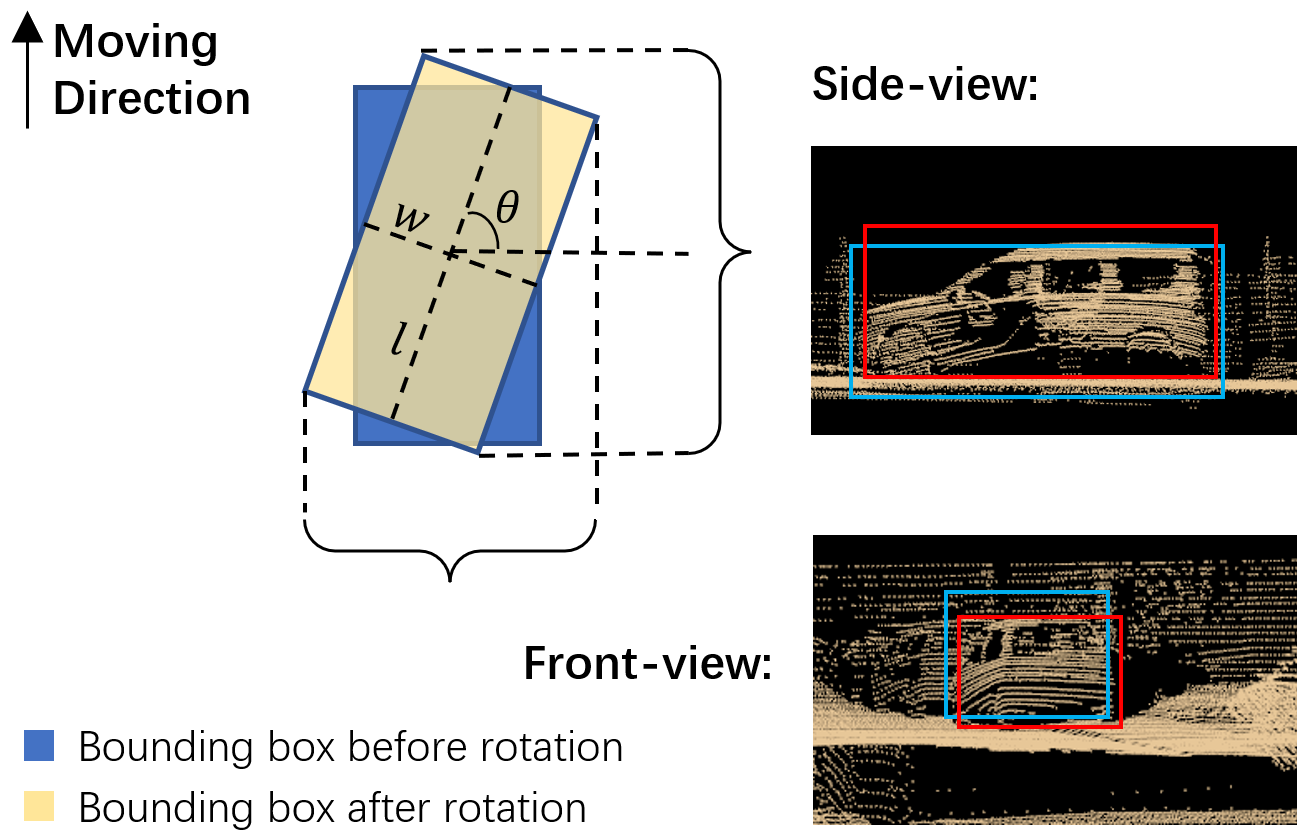} 
  \label{fig:svfv}
\end{minipage}
\hfill
\begin{minipage}[t]{0.46\textwidth}
  \centering
  \captionof{figure}{\small Performance comparison of the source-only, ROS and the best ST3D (Waymo--KITTI) models on the source domain (Waymo). The results indicate that, with the improvement of the detection ability on the target domain, the performance of ST3D models on the source domain drops significantly.} \vspace{0.05in}
  \includegraphics[width=\linewidth]{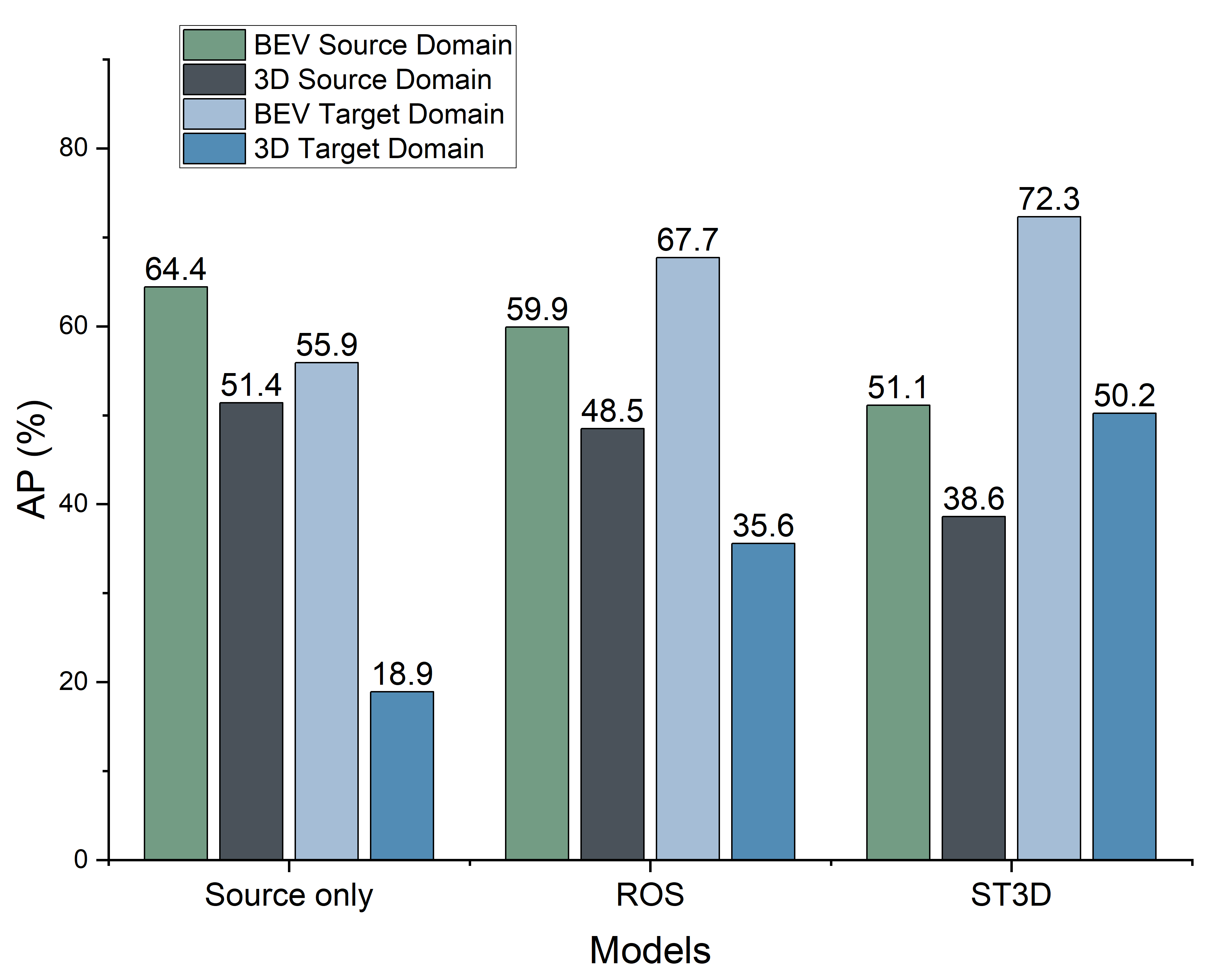}
  \label{fig:st3d}
\end{minipage}
\vspace{-0.20in}
\end{figure}

As shown in Figure \ref{fig:svfv}, we project the 3D prediction boxes of the objects into the side-view and front-view planes, \ie, the X-Z plane and the Y-Z plane. Figure \ref{fig:svfv} shows the calculation methods of these two metrics. Given a 3D bounding box with size $(2l, 2w, h)$, center $(x, y, z)$ and rotation angle $\theta$, then the projected length $l_{p}$ and width $w_{p}$ respectively to the front-view plane and side-view plane are given by
\begin{align}
	l_{p} = 2 ( w \sin\theta + l \cos\theta), \quad 
        w_{p} = 2 ( w \cos\theta + l \sin\theta).
\end{align}
Note that there is no need to project the height since the bottom side of the bounding box is parallel to the horizontal plane. We then calculate the related 2D AP with the intersection over union (IoU) thresholds at 0.7. In other words, we mark an object as being correctly detected if the IoU between the prediction box and the ground-truth box is larger than 0.7.

We focus on the performance of the \textit{Car} category -- the main focus in most existing works and datasets. KITTI evaluates three cases: Easy, Moderate, and Hard. Following~\cite{9156543}, we replace the constraints of box height with object location depth as the criteria for difficulties to better evaluate the performance on other datasets. In detail, we replace the constraints of ``larger than $40, 25, 25$ pixels'' by ``within $30, 70, 70$ meters'' for {Easy}, {Moderate}, and {Hard} difficulties.


\vspace{-0.25in}
\section{Experiments and Analysis}
\vspace{-0.05in}

\subsection{Results of LiDAR-only Methods}
We first evaluate the cross-domain performance of existing LiDAR-only methods. We summarize the results in Table~\ref{tab:LiDAR-only}. Three models are evaluated on three datasets: KITTI, Waymo, and nuScenes (Nusc). We report the results of three within-domain tasks and three cross-domain tasks. Since KITTI only provides the annotations in the front view while Waymo and nuScenes annotate the ring view point clouds, we adapt the models trained on Waymo and nuScenes and only use KITTI as the target domain. Table~\ref{tab:LiDAR-only} shows the results of the BEV and 3D AP. For the results of side-view and front-view AP, we defer the discussion in Section~\ref{ssec:additional_evaluation_metrics}.

We see that both PV-RCNN and SECOND-IoU work well on most within-domain tasks. Since nuScenes uses LiDAR sensors with fewer beams and the point cloud data of it is sparser than the others, the Nusc-Nusc task is harder and these results have already been close to the SOTA methods although they are slightly lower than the other two within-domain tasks. Based on the conclusion that both the point-based and voxel-based methods can achieve fairly good results in within-domain tasks, we see heavy drops when evaluating the same models on cross-domain tasks. When trained on nuScenes and evaluated on KITTI, Table~\ref{tab:LiDAR-only} shows that the BEV AP of PV-RCNN is even higher than within-domain results (\ie, nuScenes $\rightarrow$ nuScenes) for the Moderate and Hard difficulties, but the 3D AP drops by 3.2\%--12.7\%; and the 3D AP of SECOND-IoU drops by 22.6\%--25.4\%. Similar performance gaps are observed when the models are trained on Waymo. No matter designed with point-based or voxel-based structures, both models fail on all cross-domain tasks with large gaps compared with their within-domain performance if trained on target domains.

Similar to SECOND-IoU, TransFusion is also a voxel-based model. However, it uses a transformer decoder layer to learn from the LiDAR point cloud data and predict the bounding boxes instead of using pre-determined anchors. Furthermore, another transformer decoder is used to combine the image features with predictions from the LiDAR-only channel. We note that the first transformer decoder layer of TransFusion (TransFusion-L) can be independently used as a LiDAR-only detector. We first evaluate and compare it with the above LiDAR-only models. As shown in Table~~\ref{tab:LiDAR-only}, although TransFusion-L replaces the traditional 3D CNNs with transformer decoders, it still achieves similar results as PV-RCNN and SECOND-IoU on different cross-domain tasks.

The results show that most methods fail to obtain acceptable performance when trained and evaluated on different domains without domain transfer training. The cross-domain performance will be poorer when there is a big gap between the distributions of the source and target domains, indicating that the models have overfitted to the source domain. We hypothesize that the poor adaptation ability is not highly related to specific model types such as CNNs or transformers, but results from deeper structural problems.

\begin{table}[t] 
\caption{Performance of TransFusion with images. For within-domain tasks, TransFusion-LC (trained with LiDAR point clouds and images) achieves worse results than TransFusion-L (trained with LiDAR only). The results indicate that it is harder to adapt multi-modal methods to new domains without further training.}
\vspace{-0.08in}
\begin{center}
\setlength\tabcolsep{5pt}
\resizebox{1.0\textwidth}{!}{
\begin{tabular}{l ccc ccc}
\toprule
\multirow{2}{*}{Models} & \multicolumn{3}{c}{Nusc-Nusc} & \multicolumn{3}{c}{Nusc-KITTI} \\ 
\cmidrule(lr){2-4} \cmidrule(lr){5-7}
                        & Easy & Moderate & Hard & Easy & Moderate & Hard \\
\midrule
TransFusion-L           & 77.9 / 42.2 & 45.7 / 23.5 & 42.9 / 22.4 & 59.6 / 25.0 & 48.9 / 17.4 & 48.5 / 17.0 \\
TransFusion-LC          & 77.4 / 46.4 & 42.9 / 23.8 & 42.5 / 22.9 & 29.6 / 0.1  & 24.4 / 0.5  & 25.7 / 0.5  \\
\bottomrule
\end{tabular}
}
\label{table:TransFusion}
\end{center}
\vspace{-0.20in}
\end{table}

\begin{table*}[t]
\scriptsize
\setlength\tabcolsep{1.5pt}
\caption{Performance of ST3D from Waymo to KITTI in the format of BEV/3D AP. Different proportions of KITTI data are used for the self-training step of ST3D for comparison. When using 50\% or fewer data, we randomly sub-sample the dataset twice and report the worst and the best results of each for ST3D. In particular, as an example,  `A - Epoch 13 \& 17' means that this is the first randomization with \eg \ 50\% of KITTI, where the worst and the best results obtained at epochs 13 and 17 are shown in the middle three columns and the last three columns, respectively. The results of the source only and ROS are only shown once (in the middle three columns) since they are unique. The best results of ST3D for all three difficulties are indicated in bold.}
\vspace{-0.08in}
\begin{center}
\resizebox{1.0\textwidth}{!}{
\begin{tabular}{c |c|c|c|c||c|c|c} 
\hline
Data                 & Worst \& Best          & Easy        & Moderate    & Hard        & Easy        & Moderate    & Hard         \\ 
\hline
\multirow{3}{*}{100\%} & Source only         &   75.4 / 25.1          &     55.9 / 18.9       &      53.2 / 17.8       & -            &   -          &     -         \\ 
                        & ROS & 88.5 / 46.6   & 67.7 / 35.6   &  68.3 / 36.8  &  - & -  & -  \\
                        & ST3D (w/ ROS)         & 88.4 / 50.9 & 66.0 / 38.2 & 66.4 / 39.0 & 91.0 / 67.3 & \textbf{72.3} / 50.2 & 72.6 / 50.8  \\
                        
\hline
\multirow{2}{*}{50\%} & A - Epoch 13 \& 17 & 88.8 / 57.7 & 80.6 / 52.8 & 78.9 / 52.2 & 90.6 / 74.5 & 81.6 / 64.1 & 79.7 / 62.3  \\
                      & B - Epoch 01 \& 05 & 90.0 / 59.7 & 80.4 / 53.6 & 78.7 / 52.9 & 91.5 / 75.3 & 82.5 / 65.0 & 80.5 / 63.3             \\ 
\hline
\multirow{2}{*}{25\%} & A - Epoch 21 \& 07 & 77.7 / 27.9 & 69.6 / 28.3 & 70.0 / 28.5 & 89.7 / 60.9 & 79.1 / 54.8 & 77.1 / 53.8  \\
                      & B - Epoch 10 \& 20 & 85.3 / 46.6 & 77.3 / 42.6 & 75.7 / 43.6 & 85.4 / 66.2 & 76.4 / 59.6 & 76.3 / 59.7  \\ 
\hline
\multirow{2}{*}{10\%} & A - Epoch 17 \& 15 & 84.2 / 44.5 & 74.7 / 41.7 & 74.9 / 41.6 & 89.9 / 64.1 & 79.6 / 56.7 & 79.5 / 55.3  \\
                      & B - Epoch 02 \& 21 & 86.8 / 52.4 & 77.8 / 48.3 & 76.1 / 48.0 & \textbf{91.8} / \textbf{77.4} & 81.1 / \textbf{68.2} & \textbf{81.0} / \textbf{66.5}  \\
\hline
\end{tabular}
}
\end{center}
\label{tab:ST3D}
\vspace{-0.27in}

\end{table*}

\vspace{-0.05in}
\subsection{Results of LiDAR+RGB Methods}

We then analyze the cross-domain performance of TransFusion-LC, \ie, the multi-modal version of TransFusion taking both the LiDAR point clouds and RGB images as the input. The second decoder layer is the key point of TransFusion-LC, which helps build a soft association between the features of the point clouds and images instead of heavily relying on the calibration files. Therefore, the influence of images can be better observed without the interference of calibration files. Since OpenPCDet only supports LiDAR-only models, we use another code base, MMDetection3D \cite{mmdet3d2020}, to compare the performance of TransFusion-L and TransFusion-LC.

We must note that it is much harder to adapt multi-modal models to different domains than LiDAR-only models. Besides ensuring the preprocessing settings are the same or similar, we also need to make the images consistent. Unfortunately, existing datasets use different RGB cameras to collect the image data, which raises a challenge in making the models able to take different sizes of images as the input, during the training and evaluation stages. For example, nuScenes collects images of size 1600 $\times$ 900, while KITTI collects images of size around 1280 $\times$ 384. When adapting the models trained on nuScenes to KITTI, we can either simply pad the KITTI images to the nuScenes size during evaluation or downsample the nuScenes image size during training to keep the image size consistent. We also try to use an equally small image size of 400 $\times$ 224 or 192 $\times$ 640 to focus on the center contents of the images. Surprisingly, all methods failed to obtain reasonable results on nuScenes to KITTI tasks. 

We report the best results in Table~\ref{table:TransFusion}, with the third strategy using an image size of 400 $\times$ 224. TransFusion-LC performs similarly to TransFusion-L when trained and evaluated on nuScenes, but the performance drops heavily when evaluated on KITTI. We hypothesize that this drop comes from the large gap of images in nuScenes and KITTI. Although we have tried different methods to overcome the size difference, the pixel distribution and semantic knowledge of images are still very different in these two datasets, which makes it hard to extract useful features from KITTI images using a model trained on nuScenes. As a result, the image features become noise in the second decoder layer and lead to a heavy drop in the final detection performance. Approaches to combining the two modal data consistently and better camera-only 3D object detection methods are both necessary to reduce the noise.


\vspace{-0.05in}
\subsection{Results of Self-training Methods}


The performance of a self-training algorithm, ST3D, is analyzed. We focus on applying ST3D on PVRCNN, from Waymo to KITTI. As a self-training method, ST3D learns from the target domain without requiring ground truth annotations. Following the original work, it takes three steps in the model training process: (i) training the model normally in the source domain; (ii) using the ROS method to improve the generalization ability of the model; and (iii) training the model with the ST3D self-training algorithm by using the target data without annotations. Since self-training models are hard to converge, selecting the epoch with the best evaluation results is required \cite{9578132}. We argue that this operation should not be used; otherwise this implies the information from the annotations is used and ST3D will no longer be a self-training model. Therefore, we select the worst and the best epochs for a fairer comparison. Note that a random epoch selection may be more reasonable when applying ST3D in practice.




Table~\ref{tab:ST3D} (first row) shows that the cross-domain performance improves by a surprising degree benefiting from the additional prior knowledge about the target domain, compared with the source-only (\ie, directly adapt models from the source domain to the target domain) results. ROS improves the results by over 37\% and ST3D improves the results by over 48\% on average for the 3D AP. However, we notice two problems with ST3D. The first is that the ROS method has already boosted the performance, which means further self-training is not necessary. The second is that ST3D requires all the training data from the target domain although the annotations are not needed. This leads to the following question: Is the performance of ST3D related to the number of data samples from the target domain? We, therefore, train ST3D with fewer samples from KITTI and evaluate the models on the full KITTI dataset. We sample 50\%, 25\% and 10\% of the KITTI training data and randomize the procedure twice. We observe huge fluctuations in the results of ST3D as shown in Table~\ref{tab:ST3D}. It indicates that when trained with the full KITTI dataset, the result of the worst epoch is even lower than that of the ROS model, while the best result is around 12\% better than ROS. When trained with 25\% of KITTI, the results of the first randomization fluctuate between 27.9\% and 60.9\%, while the results of the second randomization fluctuate between 46.6\% and 66.2\%, for the 3D AP of the Easy difficulty case. When trained with just 10\% of KITTI, the results become even better with a smaller fluctuation, showing the instability of ST3D regarding different data sizes from the target domain.

It is also worth investigating whether the final ST3D model can still work well on the \textit{source} domain. In Figure~\ref{fig:st3d}, we compare the results of the source-only, ROS and the best ST3D models on Waymo. We see that with better performance on the target domain, the detection ability on the source domain becomes worse. In other words, ST3D actually shifts the knowledge domain learned by the model, rather than preserving the generalization ability. 
We thus reach three below arguments.
\begin{itemize}
\item[(I)] ST3D learns the data distribution from the target domain and its performance highly relies on the quality of the available target domain data. In other words, if the sampled 10\% of KITTI has a similar distribution to the full KITTI, ST3D can achieve good performance (\eg, 77.4\% of 3D AP in the Easy difficulty); otherwise, ST3D's performance may degrade (\eg, 27.9\% of 3D AP in the Easy difficulty). 

\item[(II)] The performance of ST3D is unstable due to its self-supervised algorithm. Even if the sampled data has a close distribution to the full target domain, it may still achieve poor results if models from bad epochs are selected.

\item[(III)] Models trained with ST3D can only work well on the target domain but cannot work well on the source domain anymore, which means the generalization ability of the models is still at a low level and the cost of generalizing the models would still be an open problem.
\end{itemize}


\begin{table*}[tbp]
\scriptsize
\setlength\tabcolsep{5.5pt}
\caption{Performance of 3D object detection models within and across multiple datasets in side-view, front-view and BEV AP with the IoU threshold set to 0.7. Results are reported in Easy / Moderate / Hard difficulties. Results show that the side-view AP is much lower than the front-view and BEV AP for most tasks, which attributes the problem to length (depth) and height errors.}
\vspace{-0.05in}
\begin{center}
\resizebox{1.0\textwidth}{!}{
\begin{tabular}{c c c c c} 
\toprule
Tasks & Metrics & PV-RCNN & SECOND-IoU & TransFusion-L \\ 
\midrule
\multirow{3}{*}{KITTI $\rightarrow$ KITTI}     
     & Side-view & 95.3 / 84.0 / 82.6 & 94.8 / 81.1 / 79.6 & 91.2 / 77.4 / 76.0 \\
     & Front-view & 98.1 / 85.0 / 84.7 & 97.7 / 82.5 / 80.5 & 95.0 / 77.7 / 76.6 \\
     & BEV & 95.0 / 81.7 / 81.4 & 94.0 / 76.5 / 76.2 & 90.9 / 73.7 / 73.0 \\
\midrule
\multirow{3}{*}{nuScenes $\rightarrow$ nuScenes}     
     & Side-view & 54.9 / 49.5 / 49.5 & 53.5 / 48.2 / 48.2 & 48.5 / 43.6 / 43.6 \\
     & Front-view & 56.1 / 50.4 / 50.4 & 53.9 / 48.4 / 48.4 & 49.3 / 45.0 / 45.0 \\
     & BEV & 57.0 / 51.7 / 51.7 & 55.0 / 49.9 / 49.9 & 54.0 / 48.8 / 48.8 \\
\midrule
\multirow{3}{*}{nuScenes $\rightarrow$ KITTI}     
     & Side-view & 82.0 / 60.9 / 61.1 & 61.9 / 39.4 / 38.1 & 57.9 / 35.9 / 36.7 \\
     & Front-view & 80.9 / 58.3 / 57.6 & 49.3 / 30.0 / 29.6 & 42.2 / 27.4 / 27.9 \\
     & BEV & 80.9 / 63.0 / 62.0 & 56.4 / 37.2 / 36.3 & 54.6 / 38.2 / 38.4 \\
\midrule
\multirow{3}{*}{Waymo $\rightarrow$ nuScenes}     
     & Side-view & 33.2 / 29.6 / 29.6 & 31.0 / 27.2 / 27.2 & 31.4 / 28.0 / 28.0 \\
     & Front-view & 34.7 / 30.5 / 30.5 & 32.6 / 28.9 / 28.9 & 34.3 / 30.2 / 30.2 \\
     & BEV & 37.0 / 32.7 / 32.7 & 32.6 / 29.2 / 29.2 & 34.9 / 31.4 / 31.4 \\
\midrule
\multirow{3}{*}{Waymo $\rightarrow$ KITTI}     
     & Side-view & 89.7 / 72.8 / 72.4 & 77.9 / 62.5 / 59.4 & 86.4 / 70.5 / 69.7 \\
     & Front-view & 72.1 / 59.4 / 60.1 & 57.2 / 46.0 / 44.5 & 78.3 / 61.8 / 63.0 \\
     & BEV & 75.4 / 55.9 / 53.2 & 66.3 / 48.1 / 45.2 & 73.6 / 55.6 / 54.7 \\

\bottomrule
\end{tabular}
}
\end{center}
\label{tab:svfv}
\vspace{-0.15in}
\end{table*}

\vspace{-0.05in}
\subsection{Analysis with Additional Evaluation Metrics}
\label{ssec:additional_evaluation_metrics}


\begin{table*}[htbp]
\scriptsize
\setlength\tabcolsep{6pt}
\caption{Performance of 3D object detection models within and across multiple datasets in different dimensions with the IoU threshold set to 0.85. We report the overlaps of the ground truth and the predictions in length (depth), width, and height in the format of Easy / Moderate / Hard difficulties.}
\vspace{-0.08in}
\begin{center}
\resizebox{1.0\textwidth}{!}{
\begin{tabular}{c c c c c} 
\toprule
Tasks & Metrics & PV-RCNN & SECOND-IoU & TransFusion-L \\ 
\midrule
\multirow{3}{*}{KITTI $\rightarrow$ KITTI}     
     & Length & 91.0 / 77.6 / 75.9 & 89.0 / 71.8 / 70.1 & 83.2 / 69.9 / 66.9 \\
     & Width  & 92.8 / 75.5 / 75.1 & 92.1 / 72.2 / 70.2 & 84.3 / 65.8 / 64.1 \\
     & Height & 94.2 / 78.5 / 78.6 & 93.2 / 77.1 / 75.7 & 88.0 / 72.0 / 70.6 \\
\midrule
\multirow{3}{*}{nuScenes $\rightarrow$ nuScenes}     
     & Length & 50.9 / 46.9 / 46.9 & 46.8 / 42.9 / 42.9 & 43.6 / 40.2 / 40.2 \\
     & Width  & 52.7 / 47.4 / 47.4 & 49.3 / 44.3 / 44.3 & 47.1 / 42.2 / 42.2 \\
     & Height & 45.9 / 40.9 / 40.9 & 43.7 / 38.9 / 38.9 & 38.2 / 33.9 / 33.9 \\
\midrule
\multirow{3}{*}{nuScenes $\rightarrow$ KITTI}     
     & Length & 57.8 / 45.7 / 45.2 & 40.1 / 28.0 / 26.5 & 28.9 / 22.8 / 23.7 \\
     & Width  & 53.4 / 42.0 / 41.5 & 18.7 / 13.7 / 14.1 & 15.6 / 12.4 / 14.0 \\
     & Height & 61.8 / 44.8 / 46.3 & 45.8 / 28.6 / 28.7 & 41.3 / 25.8 / 27.3 \\
\midrule
\multirow{3}{*}{Waymo $\rightarrow$ nuScenes}     
     & Length & 26.8 / 23.9 / 23.9 & 22.5 / 19.7 / 19.7 & 22.3 / 20.0 / 20.0 \\
     & Width  & 32.2 / 28.9 / 28.9 & 29.6 / 26.4 / 26.4 & 31.1 / 27.4 / 27.4 \\
     & Height & 27.1 / 24.2 / 24.2 & 26.5 / 23.6 / 23.6 & 26.0 / 22.9 / 22.9 \\
\midrule
\multirow{3}{*}{Waymo $\rightarrow$ KITTI}     
     & Length & 73.9 / 54.2 / 50.5 & 58.9 / 43.0 / 39.4 & 65.3 / 50.0 / 47.8 \\
     & Width  & 12.1 / 13.3 / 14.3 & 10.5 / 10.9 / 11.4 & 15.7 / 15.7 / 17.4 \\
     & Height & 89.0 / 73.2 / 74.1 & 66.9 / 55.3 / 54.2 & 86.8 / 70.7 / 72.2 \\

\bottomrule
\end{tabular}
}
\end{center}
\label{tab:eachdim}
\vspace{-0.1in}
\end{table*}

To analyze the cross-domain problem more deeply as discussed in Section~\ref{metrics}, we use the two proposed AP metrics, \ie, the side-view and the front-view AP, to evaluate the models' cross-domain performance more comprehensively.

\begin{table}[t]
\caption{The average size (meters) of 3D ground-truth bounding boxes of the five datasets and percentage differences between selected datasets.}
\vspace{-0.08in}
\label{table:size}
\setlength\tabcolsep{3pt}
\footnotesize
\centering
\resizebox{0.9\textwidth}{!}{
\begin{tabular}{lcccccccc}
\toprule
Metric & KITTI & Argoverse & nuScenes & Lyft & Waymo & W $\rightarrow$ K & N $\rightarrow$ K & W $\rightarrow$ N \\ 
\midrule
Width (m)       & 1.62 & 1.96 & 1.96 & 1.91 & 2.11 & +30.2\% & +21.0\% & +7.7\% \\
Height (m)      & 1.53 & 1.69 & 1.73 & 1.71 & 1.79 & +17.0\% & +13.1\% & +3.5\% \\
Length (m)      & 3.89 & 4.51 & 4.64 & 4.73 & 4.80 & +23.4\% & +19.3\% & +3.4\% \\
\bottomrule
\end{tabular}
}
\vspace{-0.15in}
\end{table}

We evaluate the previous models on the same tasks and report the results of the side-view, front-view, and BEV AP in Table~\ref{tab:svfv}. We see that the AP under these three metrics are similar in within-domain tasks and the cross-domain Waymo $\rightarrow$ nuScenes task; however, in the cross-domain Waymo $\rightarrow$ KITTI and nuScenes $\rightarrow$ KITTI tasks, the front-view AP is obviously lower than the side-view and BEV AP. We thus further calculate the AP of the single-dimension IoU of the length (depth), width, and height below.

Since it is easier for the single-dimension IoU to reach 0.7, we set the threshold as 0.85 instead of 0.7, and the threshold 0.85 is a more equivalent threshold against the threshold  0.7 of 2D IoUs (\ie, side-view, front-view, and BEV) for a single dimension. As shown in Table~\ref{tab:eachdim}, the length AP is much higher than the width AP for the cross-domain tasks where KITTI is the target domain, which contradicts our common sense that width might be easier to predict since it will face us in most cases and therefore has more points surrounding it. By analyzing the difference in average object size between different datasets as shown in Table~\ref{table:size}, we find that the surprising results are actually due to the degree of size difference. Specifically, for Waymo $\rightarrow$ KITTI and nuScenes $\rightarrow$ KITTI tasks, the object size differences in width (\ie, 30.2\% and 21.0\%) are bigger than that in length (\ie, 23.4\% and 19.3\%). Meanwhile, since more points surround the front-view surfaces consisting of width and height, it is easier for models to get sufficient information from the point clouds to predict the width based on the knowledge obtained from their training domains, which, together with the larger gap in width, results in bigger errors in predicting the width.

We also notice that for the cross-domain Waymo $\rightarrow$ nuScenes task in Table~\ref{tab:eachdim}, the width AP is higher than the length AP and height AP. Since the point cloud data in nuScenes is much sparser than that in Waymo, the results indicate that the sparsity of the dataset has a larger influence on the correctness of size prediction. In detail, since there are only a few points surrounding the front-view surface of the object, and the number of points in the Z-axis is limited by the LiDAR beams, the models can thus predict the width a bit better than the length and height.

\vspace{-0.0in}
\section{Conclusion}

Deep investigations on domain adaptation for 3D object detection are undertaken in this paper. Since researchers are currently focusing on achieving higher performance on a specific dataset, it is unsurprising that existing models actually overfit the training domain and cannot be directly adapted to other domains with different data distributions. It is however worth pointing out that better domain adaptation approaches are still waiting to be explored to improve the generalization ability of models instead of shifting the knowledge domain. Meanwhile, we propose two new evaluation metrics -- the side-view and the front-view AP -- to provide a more comprehensive measurement of models' cross-domain performance. By using the proposed metrics and further analyzing the performance in each dimension, we notice that the poor cross-domain performance mainly results from the width dimension when the source and target domain have similar point cloud densities, which further indicates the severe overfitting problem of existing model structures and training strategies. Our results also show that the original evaluation metrics are sometimes insufficient to analyze and guide the learning situation of models. We hope that the new side-view and front-view metrics proposed in this paper can be widely applied in the design of new 3D object detection models and the evaluation on different datasets for within-domain and cross-domain tasks.

\subsubsection{\discintname}
The authors have no competing interests to declare that are
relevant to the content of this article.
%
%
%
\bibliographystyle{splncs04}


\bibliography{mybibliography}





\end{document}